\documentclass[letterpaper, 10 pt, conference]{ieeeconf}  % Comment this line out if you need a4paper

\pdfoutput=1
\IEEEoverridecommandlockouts                              % This command is only needed if 
\usepackage[linesnumbered,ruled]{algorithm2e}
\usepackage{float}
\usepackage{graphics} % for pdf, bitmapped graphics files
\usepackage{epsfig} % for postscript graphics files
\usepackage{mathrsfs}
\usepackage{times} % assumes new font selection scheme installed
\usepackage{amsmath} % assumes amsmath package installed
\usepackage{amssymb}  % assumes amsmath package installed

\newcounter{RNum}
\renewcommand{\theRNum}{\arabic{RNum}}
\newcommand{\Remark}{\noindent\textit{\textbf{Remark}~\refstepcounter{RNum}\textbf{\theRNum}: }}
\usepackage{color}
\usepackage{color}
\usepackage[table]{xcolor}
\usepackage[T1]{fontenc}
\usepackage{bm}
\usepackage{cite}
\usepackage{hyperref}
\usepackage{float}
\usepackage[caption=false,font=footnotesize]{subfig}
\usepackage{multirow}
\hypersetup{
	colorlinks=true,
	linkcolor=blue,
	filecolor=blue,      
	urlcolor=blue,
	citecolor=cyan,
}

\title{\LARGE \bf
	Siamese Anchor Proposal Network for High-Speed Aerial Tracking} 
\author{Changhong Fu$^{1,*}$, Ziang Cao$^{2}$, Yiming Li$^{3}$, Junjie Ye$^{1}$, and Chen Feng$^{3}$% <-this % stops a space
	\thanks{$^{*}$Corresponding Author}
	\thanks{$^{1}$Changhong Fu and Junjie Ye are with the School of Mechanical Engineering, Tongji University, 201804 Shanghai, China.
		{\tt\small changhongfu@tongji.edu.cn}}%
	\thanks{$^{2}$Ziang Cao is with the School of Automotive Studies, Tongji University, 201804 Shanghai, China.}%
	\thanks{$^{3}$Yiming Li and Chen Feng are with the Tandon School of Engineering, New York University, NY 11201 New York, United States.}%}
}

\begin{document}

\maketitle
\thispagestyle{empty}
\pagestyle{empty}

%%%%%%%%%%%%%%%%%%%%%%%%%%%%%%%%%%%%%%%%%%%%%%%%%%%%%%%%%%%%%%%%
%%%%%%%%%%%%%%%%%%%%% Section 0: abstract %%%%%%%%%%%%%%%%%%%%%%
%%%%%%%%%%%%%%%%%%%%%%%%%%%%%%%%%%%%%%%%%%%%%%%%%%%%%%%%%%%%%%%%
\begin{abstract}
In the domain of visual tracking, most deep learning-based trackers highlight the accuracy but casting aside efficiency. Therefore, their real-world deployment on mobile platforms like the unmanned aerial vehicle (UAV) is impeded. In this work, a novel two-stage Siamese network-based method is proposed for aerial tracking, \textit{i.e.}, stage-1 for high-quality anchor proposal generation, stage-2 for refining the anchor proposal. Different from anchor-based methods with numerous pre-defined fixed-sized anchors, our no-prior method can 1) increase the robustness and generalization to different objects with various sizes, especially to small, occluded, and fast-moving objects, under complex scenarios in light of the adaptive anchor generation, 2) make calculation feasible due to the substantial decrease of anchor numbers. In addition, compared to anchor-free methods, our framework has better performance owing to refinement at stage-2. Comprehensive experiments on three benchmarks have proven the superior performance of our approach, with a speed of $\sim$200 frames/s.
\end{abstract}
%%%%%%%%%%%%%%%%%%%%%%%%%%%%%%%%%%%%%%%%%%%%%%%%%%%%%%%%%%%%%%%%
%%%%%%%%%%%%%%%%%%%%% Section 1: INTRODUCTION %%%%%%%%%%%%%%%%%%
%%%%%%%%%%%%%%%%%%%%%%%%%%%%%%%%%%%%%%%%%%%%%%%%%%%%%%%%%%%%%%%%
\section{Introduction}
In recent years, aerial tracking has received considerable attention because of its wide applications such as indoor obstacle avoidance~\cite{8463185}, disaster response~\cite{yuan2017aerial}, and mounting sensors~\cite{8461119}. The objective of aerial tracking is to predict the location of the object in the following frames based on its initial state. One remarkable difference between aerial tracking and general tracking is that aerial tracking requires real-time speed and low energy consumption due to the resource-constrained aerial platforms. Besides, aerial tracking suffers from various challenging scenarios introduced by unmanned aerial vehicle (UAV), \textit{e.g.}, fast motion, low resolution, and severe occlusion. Based on the aforementioned property of aerial tracking, one question is raised naturally: \textit{can we find a good balance between efficiency and robustness, and develop an efficient and effective aerial tracker?} 

In literature, aerial tracking approaches are mainly divided into two types: correlation filter-based online trackers which are CPU friendly~\cite{huang2019learning, Li_2020_CVPR, LifanICRA2020,Fu_2020_TGRS}, and deep learning-based offline trained trackers that need a high-end GPU~\cite{LiICRA2020, Fu2019IROS}. Though the former is low-cost and energy-efficient, the performance still has a clear gap compared to the latter taking advantage of the offline training. Nevertheless, the latter is suffering from low efficiency. To achieve a satisfactory balance between performance and speed, this work tries to improve deep trackers' performance while lowering their redundancy.
\begin{figure}[t]
	\centering
	% 调整比例，添加图片的相对位置
	\includegraphics[scale=0.45]{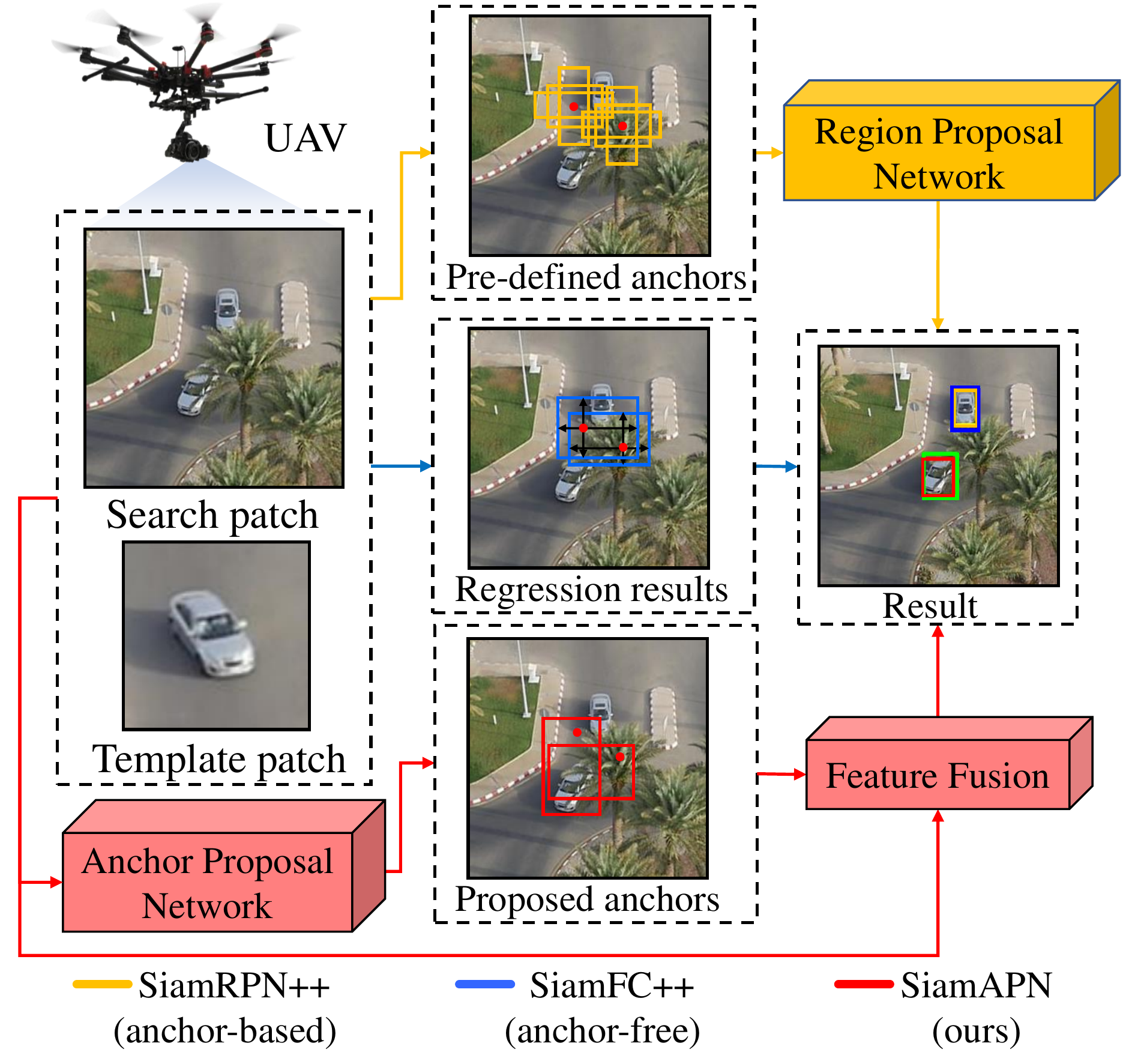}
	\caption{Comparison with the other two cutting-edge methods, \textit{i.e.}, anchors-based method (\textcolor[rgb]{1 0.75 0}{SiamRPN++}~\cite{8954116}) and anchor-free method (\textcolor[rgb]{0  0  1}{SiamFC++}~\cite{xu2020siamfc++}) on $car7$ from UAV123@10fps~\cite{Mueller2016ECCV}. The \textcolor[rgb]{0  1  0}{green} box represents the ground truth bounding box. Since the adaptive anchors generated by APN can improve the perception of \textcolor[rgb]{1  0  0}{SiamAPN}, thereby our tracker straightforward notices the occluded car. Meanwhile, the APN merely generates one no-prior anchor at each point on the feature map, reducing a large number of anchors and hyper-parameters significantly.
	}

	\label{fig:1} % 写 \label 跟在 \caption 后面，之后使用 \ref{}引用
\end{figure}

Among the deep learning-based trackers, the Siamese-based trackers play important roles in object tracking~\cite{Li_2018_CVPR,8954116,xu2020siamfc++,zhu2018distractor,9157720}. The anchor-based method was put forward firstly in SiamRPN~\cite{Li_2018_CVPR}. Recently, SiamRPN++~\cite{8954116} achieves state-of-the-art performance with more elaborate architecture and a deeper convolutional network. Since traditional anchors are pre-defined at first, RPN-based trackers are sensitive to the numbers, sizes, and aspect ratios of anchor boxes. Therefore, the hyper-parameter tuning is quite essential for those anchor-based trackers to obtain successful tracking results. To improve the generalization of trackers, the anchor-free method was proposed which directly predicts the four offsets compared to the object center~\cite{xu2020siamfc++}. Different from the two mainstream methods, our method reduces hyper-parameters hugely while utilizing the advantages of anchors by redesigning the anchor generation module.

Instead of using pre-defined anchors, this work introduces an anchor proposal network (APN) to generate one adaptive anchor for each point on feature maps. Since the adaptive anchors can adapt to the various objects, the negative samples in classification are decreased. Therefore, the APN alleviates the imbalanced samples during offline training, increasing the utilization of anchors while reducing the computation and boosting the tracking efficiency. The comparison of anchor-based~\cite{8954116}, anchor-free~\cite{xu2020siamfc++}, and our method is shown in Fig.~\ref{fig:1}. When the object is fully or partially occluded, the information of objects is highly decreased and similar objects will become more obvious on the feature map, which could induce the trackers to make false judgments. By virtue of the adaptive anchor proposal generation, our method can find out the areas where the object is most likely to exist, making light work of the classification. Please note that since the adaptive anchors can adapt to the various objects which makes the regression based on anchors become more difficult, the feature fusion network is introduced for integrating the location information of anchors to eliminate the influence. Eventually, relying on the proposed strategy, the tracking performance is raised significantly. Extensive experiments demonstrate that SiamAPN maintains the robustness of the proposed approaches in other UAV-specific challenges, especially in low resolution and fast motion.

%After stage-1 for adaptive anchor proposal generation, stage-2 will refine the proposed anchors based on the fused feature map, which fuses a deeper layer with the anchor proposal layer to raise the semantic information along with the anchor proposal knowledge.

The contributions of this work are as follows:
\begin{itemize}

	\item A novel no-prior anchor proposal network (APN) is developed to enable adaptive and lightweight anchor generation, significantly improving the adaptivity and generality of SiamAPN in aerial scenarios.
	\item An elaborate feature fusion network is proposed to fuse the location information of anchors, further enhancing the semantic knowledge of feature maps and object perception of SiamAPN.
	\item The proposed method, Siamese anchor proposal network (SiamAPN), has shown competitive performance on three challenging aerial tracking benchmarks with a promising speed of $\sim$200 FPS.

\end{itemize}
%%%%%%%%%%%%%%%%%%%%%%%%%%%%%%%%%%%%%%%%%%%%%%%%%%%%%%%%%%%%%%%%
%%%%%%%%%%%%%%%%%%%%% Section 2: RELATED WORK %%%%%%%%%%%%%%%%%%
%%%%%%%%%%%%%%%%%%%%%%%%%%%%%%%%%%%%%%%%%%%%%%%%%%%%%%%%%%%%%%%%
\section{Related Works}\label{sec:RELATEDWORK}
%\subsection{Real-time tracking for UAV}
Generally, as a result of its high-speed, discriminative correlation filter (DCF)-based trackers have been widely adopted on UAV tracking. The DCF-based approach took the object tracking algorithms to a new level, greatly promoted the robustness and accuracy with satisfying speed~\cite{henriques2014high,kiani2017learning,Fu_2020_TGRS,Li_2020_CVPR,9196530,Li_2020_ICRA,9340954,Fu_2020TGRS_DRCF}. In recent years, Siamese-based trackers have become attractive due to their state-of-the-art performance. An efficient Siamese-based tracker is also another promising choice for UAV tracking.

%\subsection{Siamese-based methods for UAV object tracking}
With the development of the convolutional neural network (CNN), the advantage of deep-learning is magnified. Meanwhile, the Siamese-based network also shows its potential. After the SINT~\cite{tao2016sint} transferred the tracking task into matching the patch, SiamFC~\cite{bertinettofully} proposed an end-to-end method of tracking by employing a fully-convolutional neutral network to learn similarity. Notwithstanding, their performance is still suboptimal. Therefore, DSiam~\cite{8237458} applied the target and background transformation method. Regarding visual tracking as a classification\&regression task, SiamRPN introduced a region proposal network (RPN) into the Siamese-based framework to improve accuracy and robustness. To overcome the influence of distractors during tracking, DaSiamRPN~\cite{zhu2018distractor} proposes a method of utilizing the information of background. SiamRPN++~\cite{8954116} achieves state-of-the-art performance by stacking the RPN module and applying Resnet~\cite{7780459} as the backbone. There is no doubt that the anchor-based method promotes the development of community significantly. Nevertheless, the brought hyper-parameters impede the generality of those trackers. Besides, due to the movement of the anchor's center point caused by regressing, it also brings errors to classification results. To handle those problems, anchor-free trackers are proposed, \textit{e.g.}, SiamFC++~\cite{xu2020siamfc++}, SiamCAR~\cite{9157720}. They eliminated the above classification errors by fixing the center point and regressing the four offsets. 

Different from the two mainstream methods, we apply a novel two-stage method to handle shortcomings of the anchor-based above. Meanwhile, SiamAPN utilizes the feature fusion network to integrate the location information of anchors for maintaining the performance of stage-2. Experiments on three UAV benchmarks demonstrate the competitive performance of SiamAPN with high-speed.
\begin{figure*}[t]
	\centering
	% 调整比例，添加图片的相对位置
	%\includegraphics[scale=0.5]{images/1.pdf}
	\scalebox{1}[1]{\includegraphics[width=1\textwidth]{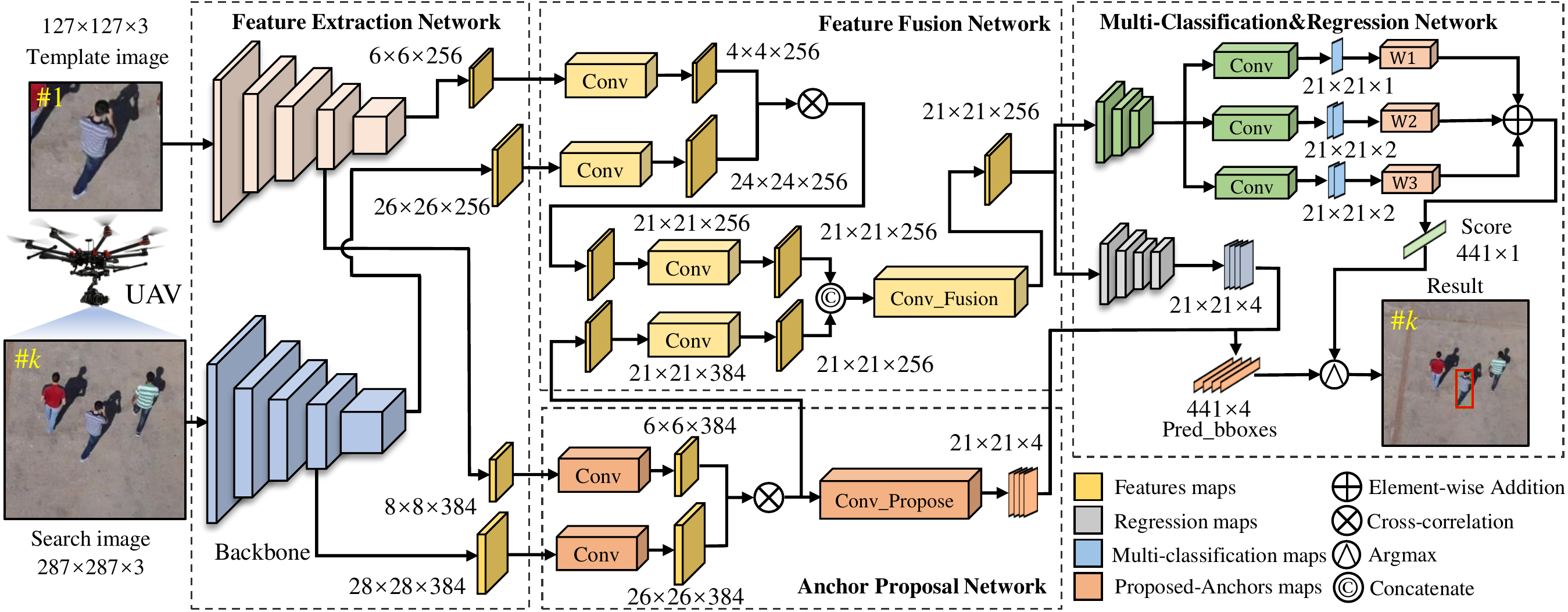}}
	%\scalebox{0.5}[0.5]{\includegraphics[trim={0 75 0 55},clip]{images/1.pdf}}
	% 写标题
	\caption{The overview of the SiamAPN tracker. It composes of four subnetworks, \textit{i.e.}, feature extraction network, feature fusion network, anchor proposal network (APN), and muti-classification\&regression network.
	}
	\label{fig:main} % 写 \label 跟在 \caption 后面，之后使用 \ref{}引用
\end{figure*}

%%%%%%%%%%%%%%%%%%%%%%%%%%%%%%%%%%%%%%%%%%%%%%%%%%%%%%%%%%%%%%%%
%%%%%%%%%%%%%%%%%%%%% Section 3: Proposed method %%%%%%%%%%%%%%%%%%
%%%%%%%%%%%%%%%%%%%%%%%%%%%%%%%%%%%%%%%%%%%%%%%%%%%%%%%%%%%%%%%%
%证明featurefusion的必要性   消融实验 证明在visual trackingdataset训练uav数据集测试必要性

\section{Proposed Method}\label{sec:proposed}
In this section, the proposed SiamAPN is introduced in detail. As shown in Fig. \ref{fig:main}, SiamAPN consists of four subnetworks, respectively the feature extraction network, APN, feature fusion network, and multi-classification\&regression network. 
\subsection{Feature extraction network}
The feature extraction network consists of a Siamese network, which has two share-architecture branches, \textit{i.e.}, the template branch and the search branch. The inputs of these two branches are respectively a template image (denoted as $ z $) and a search image (denoted as $ x $). Undergoing five $conv$ blocks, the outputs corresponding to the inputs are $\varphi_5(x)$ and $\varphi_5(z)$, each with 256 channels. To generate adaptive anchors, the feature maps extracted by the $conv4$ block, \textit{i.e.}, $\varphi_4(x)$ and $\varphi_4(z)$ with 384 channels, are exploited.

\subsection{Anchor proposal network}
For the sake of handling the special challenges in aerial tracking without sacrificing efficiency, we introduce adaptive anchors. As illustrated in Fig.~\ref{fig:main} the APN utilizes the output of the penultimate $conv$ layer ($\varphi_4(x)$, $\varphi_4(z)$) for anchor proposal. $\varphi_4(x)$ and $\varphi_4(z)$ are convoluted with a $3 \times 3$ kernel for adjusting the features from backbone. Afterward, inspired by \cite{8954116}, a depth-wise cross-correlation layer is cooperated to produce the similarity map as:
\begin{equation}
R_{1}=\mathcal{F}_{1}(\varphi_4(x))\star\mathcal{F}_{2}(\varphi_4(z))~,
\end{equation}
where $\star$ denotes the depth-wise cross-correlation operation and $\mathcal{F}_{1}(\cdot),~\mathcal{F}_{2}(\cdot)$ represent different convolution operation. The response map $R_{1}$ has the same channels as $\varphi_4(x)$ which contains the information of object.

\begin{figure}[b]
	\centering
	% 调整比例，添加图片的相对位置
	\includegraphics[scale=0.48]{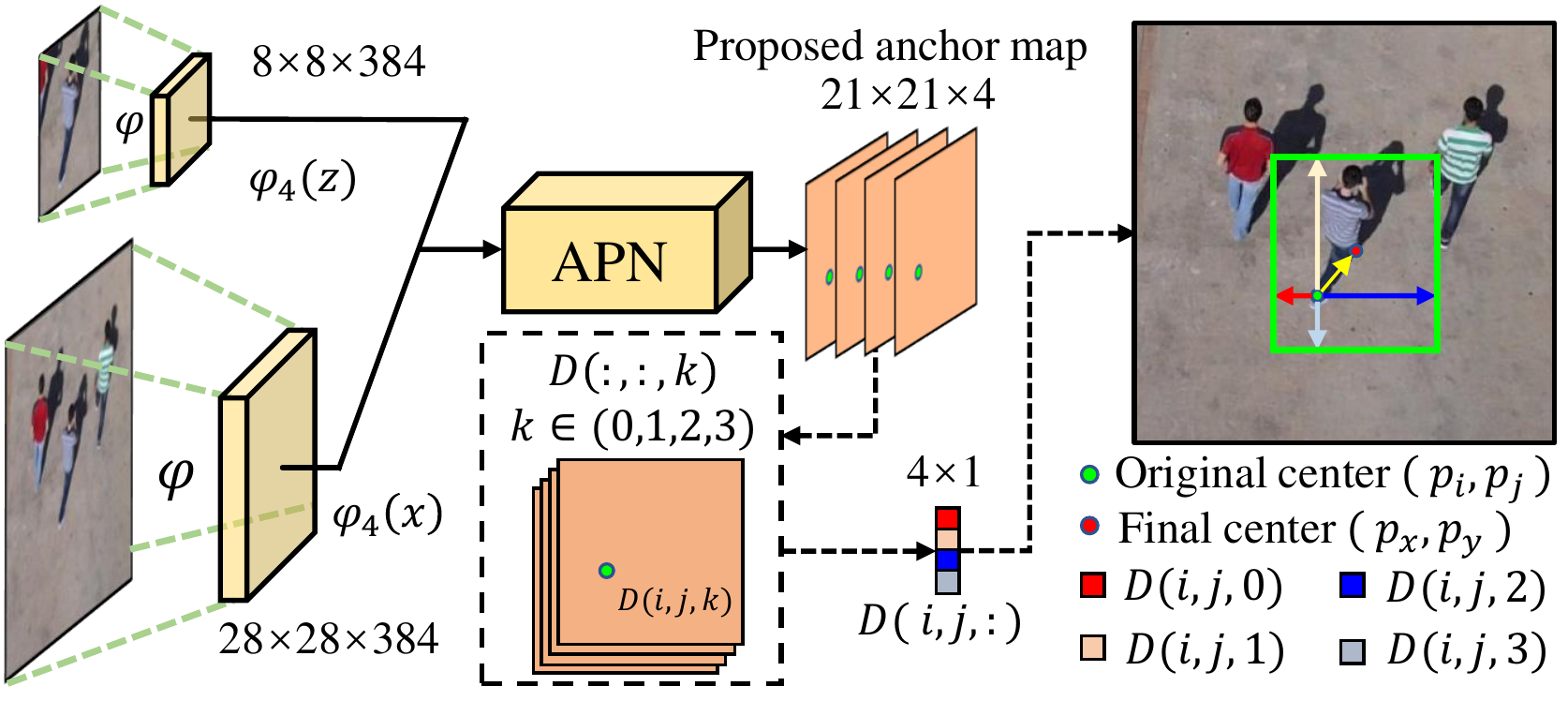}

	%\scalebox{0.5}[0.5]{\includegraphics[trim={0 75 0 55},clip]{images/1.pdf}}
	% 写标题
	\caption{Visualization of anchor proposing workflow. Each point on proposed anchor map (denoted as a \textcolor[rgb]{0  1  0}{green} point) corresponds to a certain point on the search patch $x$. By calculating four offsets, the anchors can actively adapt to the appearance of different objects. 
	}
	\label{fig:main1} % 写 \label 跟在 \caption 后面，之后使用 \ref{}引用
\end{figure}

Undergoing a convolution operation, the adaptive anchors are obtained. Note that the APN merely produces one anchor for each point in the similarity map. Each location on the proposed anchor map $D(i,j,:)$ can be mapped to the search patch. For instance, the position ($i, j$) on the proposed anchor map corresponds to the location ($p_{i}, p_{i}$), which is the center of the receptive field of ($ i, j$):
\begin{equation}
\begin{split}
p_{i}= \dfrac{w_{s}}{2}+\left(i-\dfrac{w}{2}\right)\times s~,\\
p_{j}= \dfrac{h_{s}}{2}+\left(j-\dfrac{h}{2}\right)\times s~,
\end{split}
\end{equation}
where $w_{s}$ and $h_{s}$ denote the width and height of $x$, $s$ presents the total stride of the network, and $w,~h$ are the width and height of feature maps respectively.
As illustrated in Fig.~\ref{fig:main1}, the proposed anchor map $D(i,j,:)$ corresponds to the distance between the point $p~(p_{i}, p_{j})$ and proposed anchors. 

\Remark In the anchor-based method, the anchors whose intersection over union (IoU) with ground truth are within a certain range are ignored in classification. Uniquely, since APN gives anchors adaptive characteristics, most anchors are put to good use including those ignored samples in the anchor-based method. Moreover, the experiments demonstrate the effectiveness of APN. 

Denoting $g=(g_{x1},~g_{y1},~g_{x2},~g_{y2})$ the left-top and right-bottom corners of the ground truth bounding box, the regression label of the proposed anchors $\widetilde{x}_{(i,j)}$ can be calculated by:
\begin{equation}
\begin{split}
	\widetilde{x}^{0}_{(i,j)}=p_{i}-g_{x1}~,~ \widetilde{x}^{1}_{(i,j)}=p_{j}-g_{y1}~,\\
	\widetilde{x}^{2}_{(i,j)}=g_{x2}-p_{i}~,~ \widetilde{x}^{3}_{(i,j)}=g_{y2}-p_{j}~.
\end{split}
\end{equation}

Apparently, the points far away from the
center of the target tend to produce imprecise bounding boxes, which degrade the performance of the tracking system. To remove these outliers, 
a quality weight, \textit{i.e.}, $\mathcal{W}(\cdot)$, is incorporated to the regression loss as:
\begin{equation}
	\mathcal{L}_{apn}= \sum\nolimits_{i,j}\mathcal{W}(p_{i},p_{j})L_{1}(D(i,j,:),\widetilde{x}^{n}_{(i,j)})~,
\end{equation}
where $L_{1}$ is \textrm{L1loss} and $D(i,j,:)$ represents the proposed anchor map. Besides, $\mathcal{W}$ is obtained as follows:
\begin{equation}
\mathcal{W}(i,j)=
\left\{
\begin{array}{ll}
1, &  (i,j)~\mathrm{in}~\mathbf{R}_{gtc}\\
0, & else,\\
\end{array}
\right.
\end{equation}
where $\mathbf{R}_{gtc}$ has the same center and aspect ratio as the ground truth but whose area is larger than the ground truth.

By regressing four offsets, the center and size of anchors can adapt to the variable appearance of different objects, improving the performance. Please note that albeit the generation of anchors is similar to the anchor-free method, the purpose of the adaptive anchors focuses on providing more effective location information for SiamAPN. Attributing to the improved perception brought by APN, our SiamAPN tracker can distinguish the objects from complex conditions, especially in fast motion, low-resolution, and occlusion.

\Remark Different from pre-defined anchors, the proposed anchors can adapt to the various objects during aerial tracking, especially in low resolution, fast motion, and occlusion. Moreover, by virtue of the semantic location information of adaptive anchors, SiamAPN can raise the performance of stage-2 eventually compared with the anchor-free method. 

\subsection{Feature fusion network}
After obtaining the proposed anchors, each point $(i, j)$ on the response map $D(i,j,:)$ has an adaptive anchor to locate the object coarsely, decreasing the complexity of classification. Nevertheless, the movement of the adaptive anchors brings difficulties to regression. To eliminate this drawback, the feature fusion network is created to bridge the gap between APN and the classification\&regression network. By applying the feature maps ($\varphi_5(x)$, $\varphi_5(z)$) produced by feature extraction networks, the second similarity map $R_{2}$ can be calculated by:
\begin{equation}
	R_{2}=\mathcal{F}_{3}(\varphi_5(x))\star\mathcal{F}_{4}(\varphi_5(z))~.
\end{equation} 

Some previous methods~\cite{9157720} take advantage of fusing both low-level and high-level features to improve tracking accuracy. Differently, we fuse the feature maps from APN and feature extraction network, \textit{i.e.}, $R_{3}$. Specifically, $R_{3}$ can be performed via the channel-wise concatenation and convolutional operation as follows:
\begin{equation}
	R_{3}=Cat(\mathcal{F}_{5}(R_{2}),\mathcal{F}_{6}(R_{1}))~,
\end{equation} 
\noindent where $R_{3}$ has $2\times 256$ channels. To make preparation for multi-classification\&regression, the response map $R_{3}$ is convoluted with 1$\times$1 kernel to reduce its channels to 256 (denoted as $R_{3}^{*}$), decreasing the computational load.

\Remark Since the proposed anchor maps represent the location information of anchors which contains significant information for robust tracking, the feature fusion network is constructed for integrating the internal information with high-level appearance features, enriching the semantic knowledge of the feature map.
\subsection{Multi-classification\&regression network}

After obtaining the proposed anchor map $D(i,j,:)$, each point has a corresponding anchor. To further promote the performance of classifying, multi-classification structure is utilized. The first branch aims to determine the anchor having the largest IoU with ground truth. The second branch concentrates on selecting the points on $D(i,j,:)$ that fall in the ground truth box. The last branch is design to consider the center distance between each corresponding point and ground truth inspired by~\cite{9157720}.

As illustrated in Fig. \ref{fig:main}, the classification branch outputs three classification maps ($H_{w\times h\times 2}^{cls1},~H_{w\times h\times 2}^{cls2}$, and $H_{w\times h\times 1}^{cls3}$). By integrating the three classification branches, the overall classification loss function can be calculated as:
\begin{equation}
\mathcal{L}_{cls}=\lambda_{cls1}\mathcal{L}_{cls1}+\lambda_{cls2}\mathcal{L}_{cls1}+\lambda_{cls3}\mathcal{L}_{cls2}~,
\end{equation}
where $\mathcal{L}_{cls1}$ is the cross-entropy loss functions and $\mathcal{L}_{cls2}$ represents the binary
cross entropy loss. $\lambda_{cls1}, \lambda_{cls2}, \lambda_{cls3}$ are the coefficients to weight these three branches. Note that each point in $H_{w\times h\times 2}^{cls1}$ contains 2D vector represents the quality evaluation of the corresponding anchor. While each vector on $H_{w\times h\times 2}^{cls2}$ reflect the classification score of the corresponding point. By making the classification more specified, SiamAPN can maintain promising classification performance under complex conditions.

As for the regression branch, it merely outputs one regression feature map, \textit{i.e.}, $H_{w\times h\times 4}^{loc}$. The regression label (denoted as $\widetilde{r}_{(i,j)}$) uses the 4D vectors ($\widetilde{r}^{0}_{(i,j)}, \widetilde{r}^{1}_{(i,j)}, \widetilde{r}^{2}_{(i,j)}, \widetilde{r}^{3}_{(i,j)}$) to represent the target of regression. Since the center point, width and height of the ground truth
boxes, \textit{i.e.}, $g_{x}, g_{y}, g_{w}, g_{h}$, can be calculated from $g$ respectively, the distance is obtained as:
\begin{equation}
\begin{split}
&\widetilde{r}^{0}_{(i,j)}=\dfrac{g_{x}-p_{x}}{p_{w}}~,~~ \widetilde{r}^{1}_{(i,j)}=\dfrac{g_{y}-p_{y}}{p_{h}}~,\\
&\widetilde{r}^{2}_{(i,j)}=ln\dfrac{g_{w}}{p_{w}}~,~~~~~ \widetilde{r}^{3}_{(i,j)}=ln\dfrac{g_{h}}{p_{h}}~,
\end{split}
\end{equation}
where $p'=(p_{x}, p_{y}, p_{w}, p_{h})$ represents center point and shape of the proposed anchor boxes which can be calculated by:
\begin{equation}
\begin{split}
p'=\mathcal{G}(D(i,j,:), p)~,
\end{split}
\end{equation}
where $\mathcal{G}$ represents the transformation of calculating the proposed anchor from the offset $D(i,j,:)$ and original center $p$. Putting the advantage of different loss functions into consideration, we adopt a smooth L1loss and IoUloss for obtaining robust regression. Consequently, the regression loss is computed as:
\begin{equation}
\begin{split}
\mathcal{L}_{loc}= \lambda_{loc1}L_{IOU}(\mathcal{T}(H^{loc}(i, j,:),p'), g)+\\
\lambda_{loc2}L1(H^{loc}(i, j,:), \widetilde{r}_{(i,j)}) ~,
\end{split}
\end{equation}
where $\mathcal{T}$ aims to transfer the $H^{loc}(i, j,:)$ to predicted boxes for calculating \textit{IoUloss}. 

Put all loss functions together, the overall loss function is as follows:
\begin{equation}
	\mathcal{L}=\mathcal{L}_{apn}+\lambda_{1}\mathcal{L}_{cls}+\lambda_{2}\mathcal{L}_{loc}~,
\end{equation}
where $\lambda_{1}, \lambda_{2}$ are the coefficients to weight the classification loss and regression loss.
\begin{figure*}[t]
	\centering	
	\includegraphics[width=0.32\linewidth]{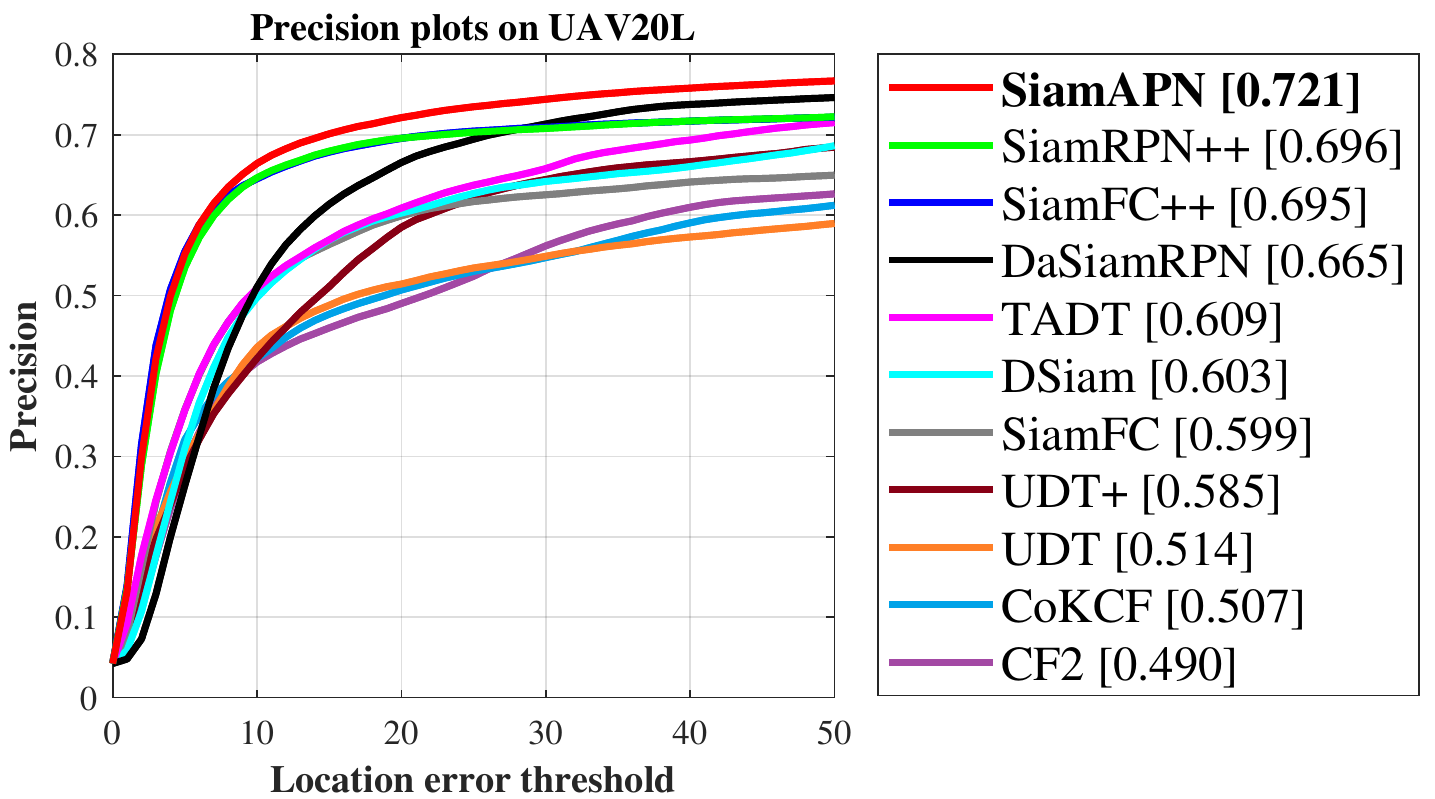}
	\includegraphics[width=0.32\linewidth]{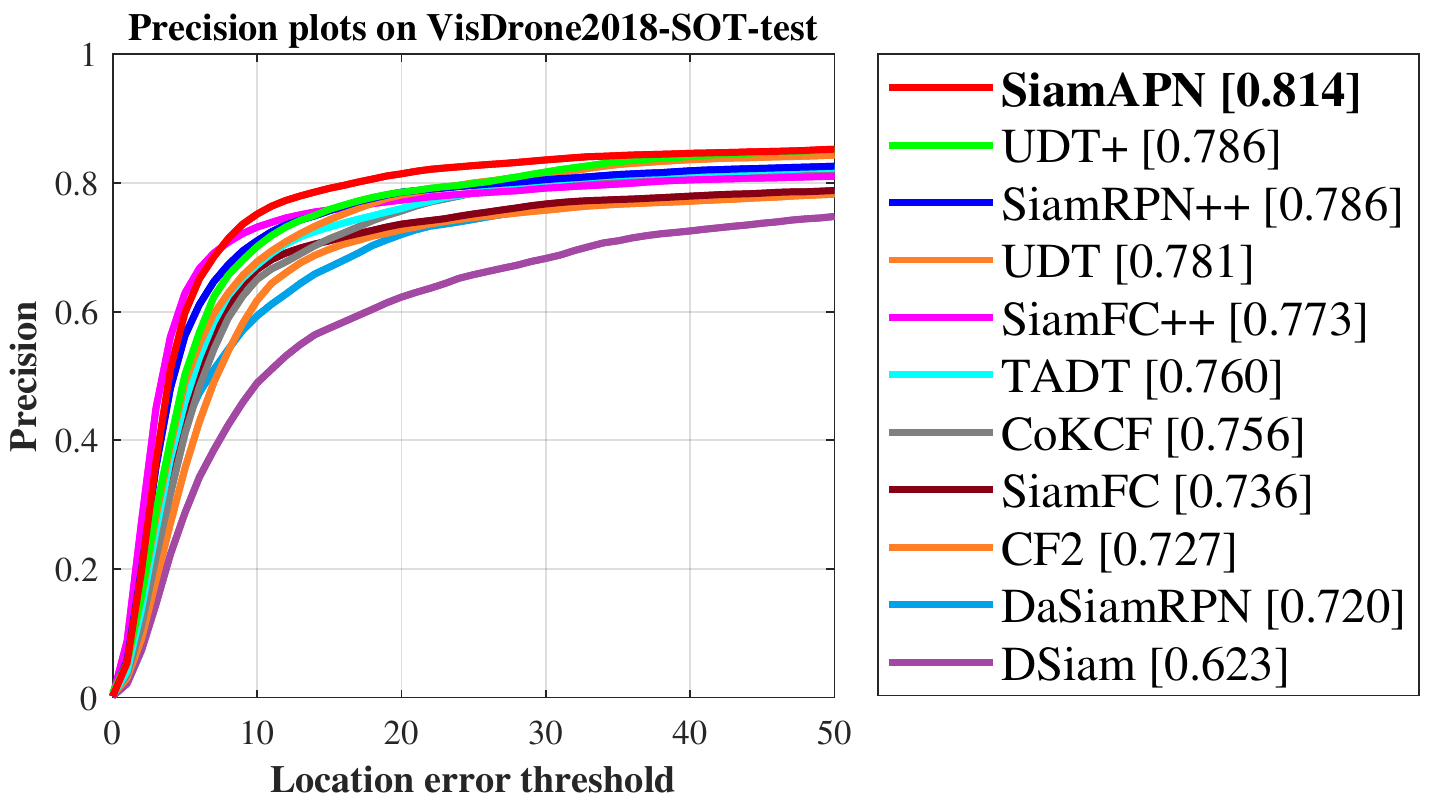}
	\includegraphics[width=0.32\linewidth]{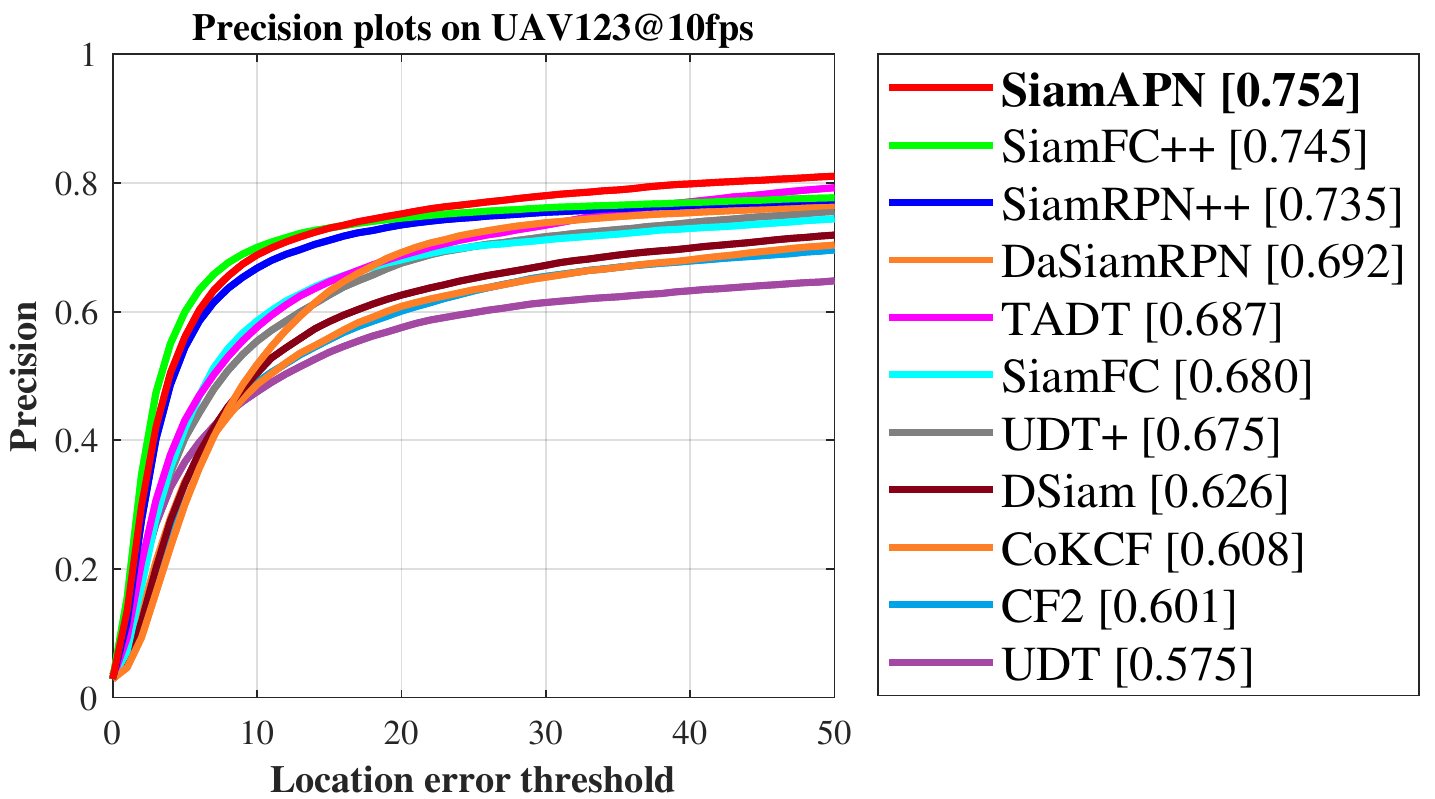}
	\subfloat[Results on UAV20L]
	{\includegraphics[width=0.32\linewidth]{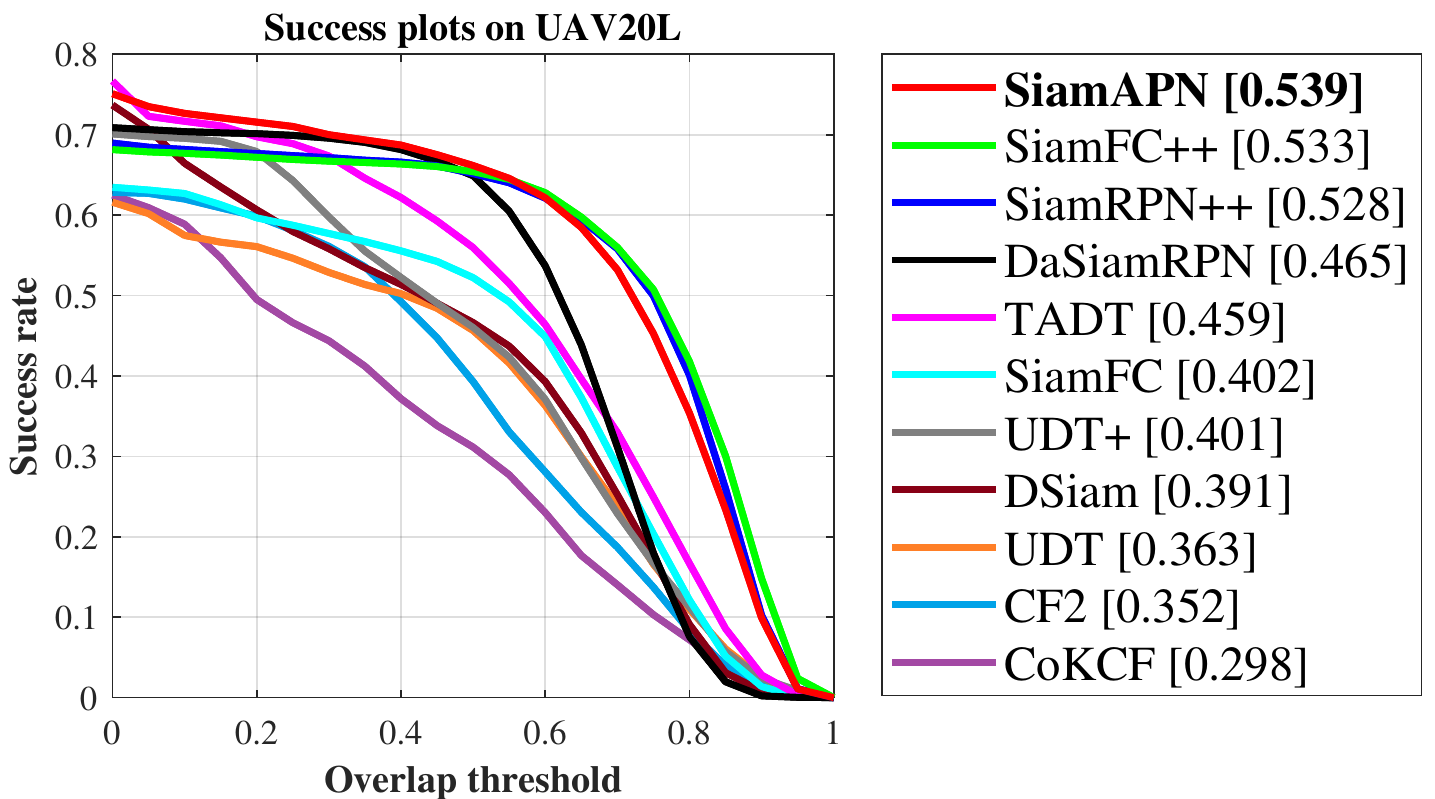}}
	\hspace{-1pt}
	\subfloat[Results on VisDrone2018-SOT-test]
	{\includegraphics[width=0.32\linewidth]{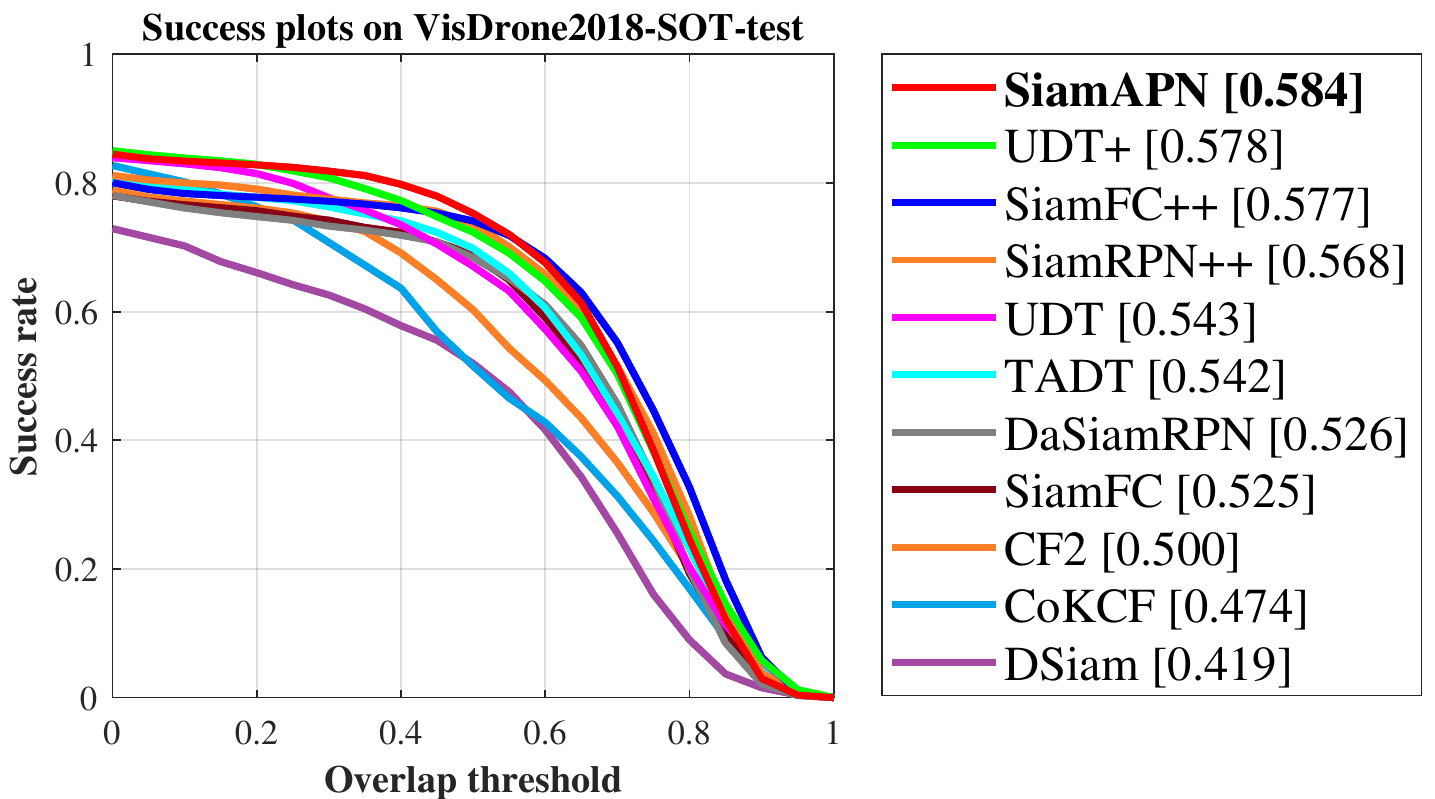}}
	\hspace{1.2pt}
	\subfloat[Results on UAV123@10fps]
	{\includegraphics[width=0.32\linewidth]{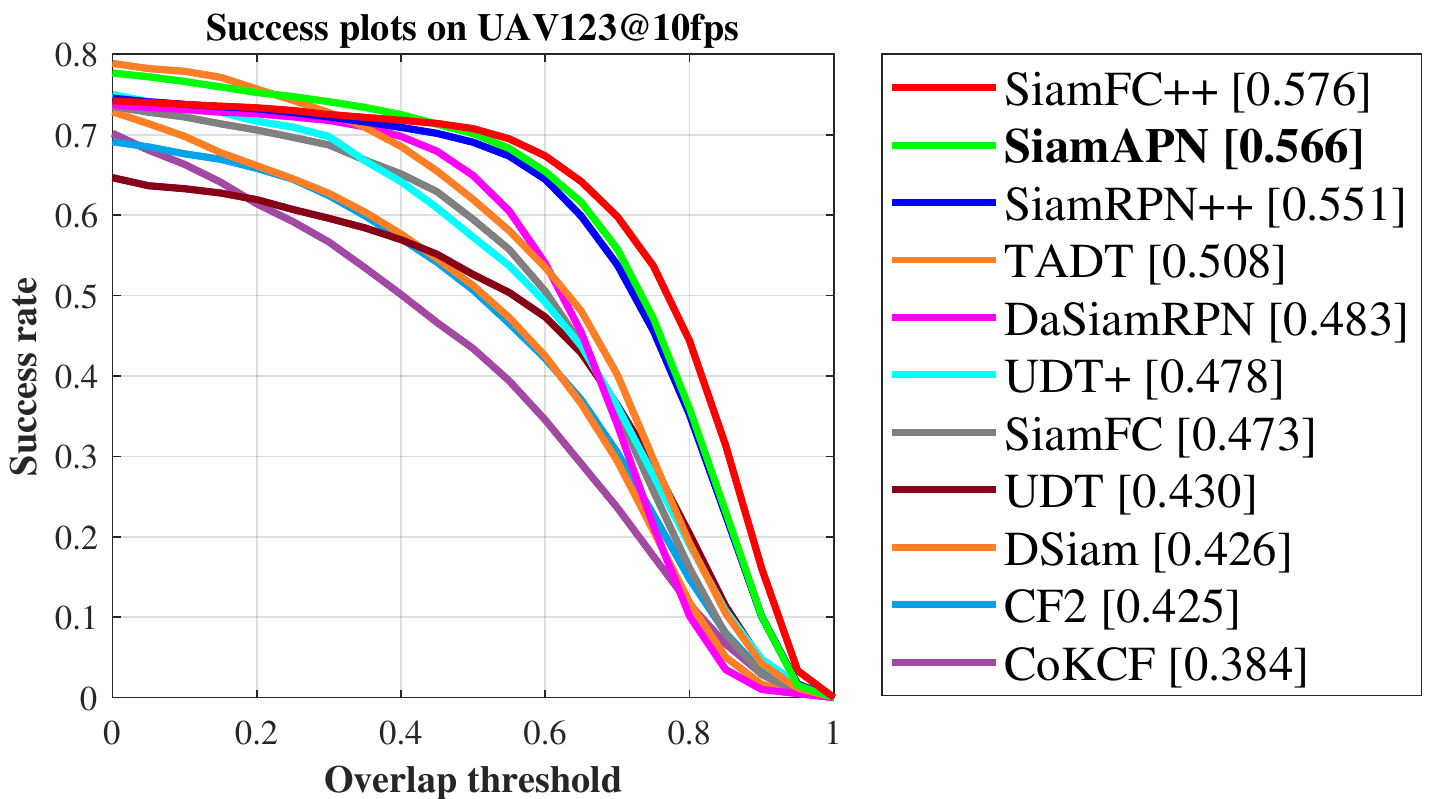}}
	\caption{Overall performance of all trackers on (a) UAV20L, (b) VisDrone2018-SOT-test, and (c) UAV123@10fps. The experimental results demonstrate that our method yields superior performance on all benchmarks.}

	\label{fig:overperform}
\end{figure*}

%%%%%%%%%%%%%%%%%%%%%%%%%%%%%%%%%%%%%%%%%%%%%%%%%%%%%%%%%%%%%%%%
%%%%%%%%%%%%%%%%%%%%% Section 4: EXPERIMENT %%%%%%%%%%%%%%%%%%%%
%%%%%%%%%%%%%%%%%%%%%%%%%%%%%%%%%%%%%%%%%%%%%%%%%%%%%%%%%%%%%%%%

\section{Experiments}\label{sec:EXPERIMENT}

In this section, the proposed method is comprehensively evaluated on three well-known authoritative UAV tracking benchmarks, \textit{i.e.},~VisDrone2018-SOT-test~\cite{wen2018visdrone}, UAV20L~\cite{Mueller2016ECCV}, and UAV123@10fps~\cite{Mueller2016ECCV}. 10 deep-learning based trackers are involved in the evaluation. Classifying by tracking speed, they can be divided as real-time trackers, \textit{i.e.}, SiamRPN++~\cite{8954116}, DaSiamRPN~\cite{zhu2018distractor}, SiamFC++~\cite{xu2020siamfc++}, SiamFC~\cite{bertinettofully}, UDT~\cite{Wang_2019_Unsupervised}, UDT+~\cite{Wang_2019_Unsupervised}, TADT~\cite{Xin2019CVPR}, DSiam~\cite{guo2017learning}, CoKCF~\cite{zhang2017robust}, and non-real-time tracker, \textit{i.e.},  CF2~\cite{Ma-ICCV-2015}. 

\Remark For the sake of fairness, all trackers are equipped with the same backbone, \textit{i.e.}, AlexNet~\cite{krizhevsky2012imagenet}, which is pre-trained on ImageNet~\cite{article}. Note that, all parameters of those trackers are consistent with their official paper, respectively.  

\subsection{Implementation details}
\label{subsec:EvaCri}

SiamAPN adopts AlexNet as the backbone with the parameters of the first two convolution layers frozen and only fine-tune the last three convolution blocks. There are a total of 50 epochs. For the first 10 epochs, the parameters of the feature extraction network are frozen to train the APN, feature fusion network, and multi-classification\&regression network. For the last 40 epoch, the whole network is end-to-end trained with a learning rate decayed from $0.005$ to $0.0005$ in log space. Besides, our network is trained with SGD with a minibatch of $124$ pairs. Besides, the momentum of $0.9$ is used. During the process of training, $\lambda_{cls1}$ is set to 1.2 while other parameters, \textit{i.e.}, $\lambda_{cls2}$, $\lambda_{cls3}$, $\lambda_{loc1}$, and $\lambda_{loc2}$, are set to 1. As for the input size of the template patch and search patch, they are set to $127 \times 127$ pixels and $287 \times 287$ pixels, respectively. Images from COCO~\cite{lin2014microsoft}, ImageNet VID~\cite{russakovsky2015imagenet}, GOT-10K~\cite{huang2019got} and Youtube-BB~\cite{real2017youtube} are extracted to train our tracker on a PC with an Intel i9-9920X CPU, a 32GB RAM, and an NVIDIA TITAN RTX GPU. The tracking code is available at \url{https://github.com/vision4robotics/SiamAPN}.

\subsection{Evaluation metrics}
The experiments are based on one-pass evaluation (OPE)~\cite{6619156}, \textit{i.e.}, precision and success rate. The success rate is measured as IoU. The success plot represents the percentage of the frames whose IoU is larger than a preset threshold. The area-under-the-curve (AUC) on the success plot is utilized for ranking. Besides, the precision is measured by the center location error (CLE) between the estimated bounding box and the ground truth bounding box. The precision plot shows the percentage of scenarios whose distance between the estimated bounding box and ground truth one is smaller than different thresholds and the score at 20 pixels is used for ranking.

\begin{figure*}[t]
	\centering	
	
	\subfloat[Fast motion on UAV123@10fps (left) and UAV20L (right)]
	{\includegraphics[width=0.245\linewidth]{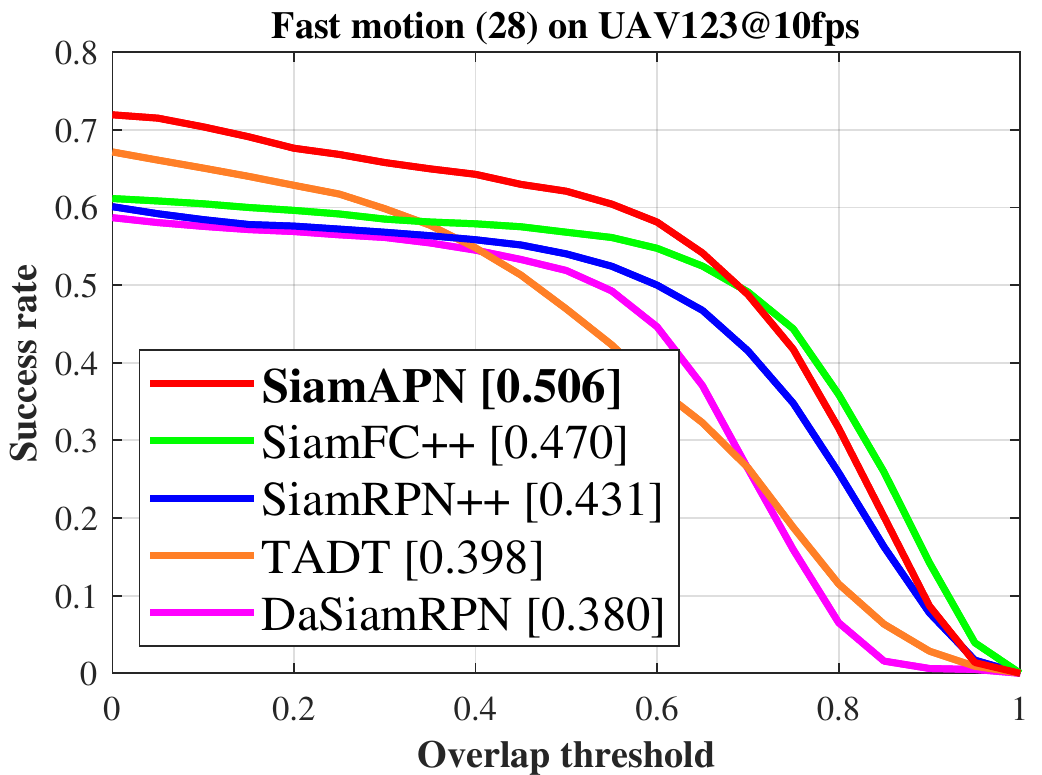}
	\includegraphics[width=0.245\linewidth]{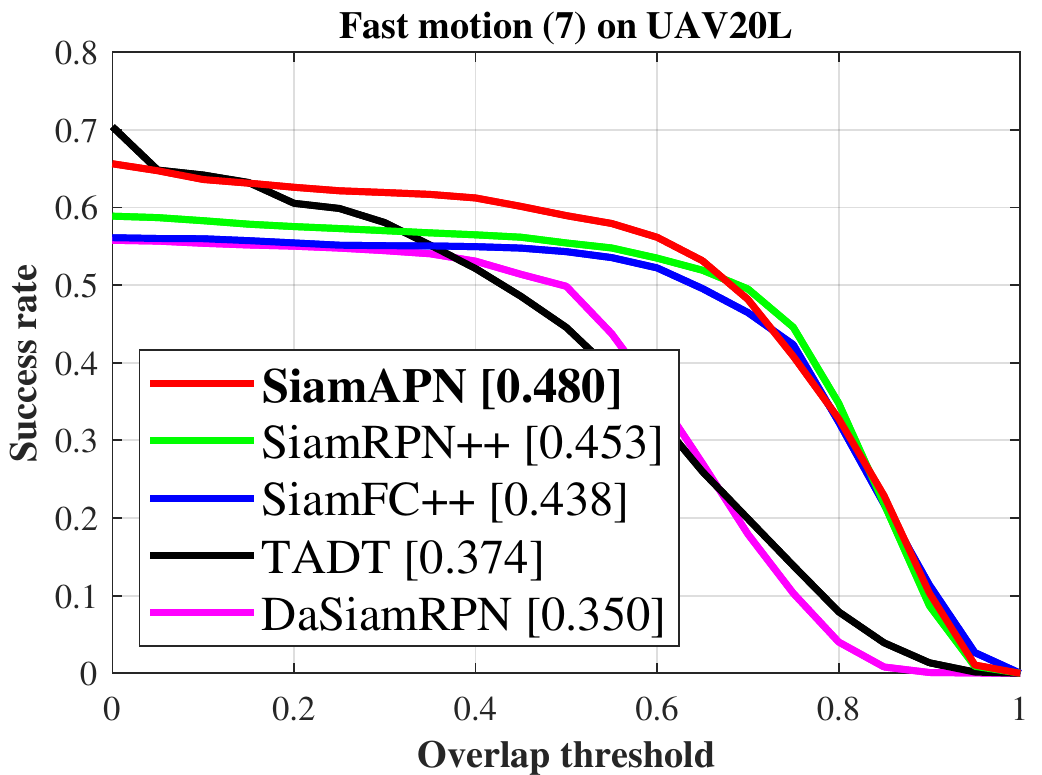}
	\label{fig:a}}
	\subfloat[Low resolution on UAV123@10fps (left) and UAV20L (right)]
    {\includegraphics[width=0.245\linewidth]{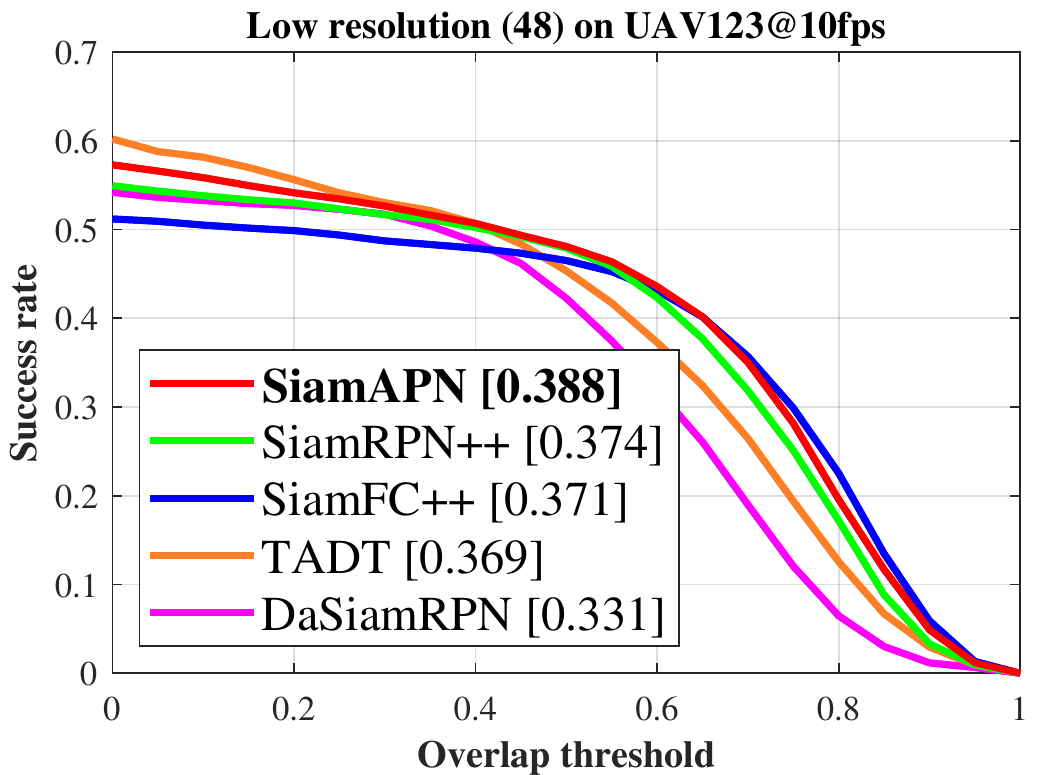}
    \includegraphics[width=0.245\linewidth]{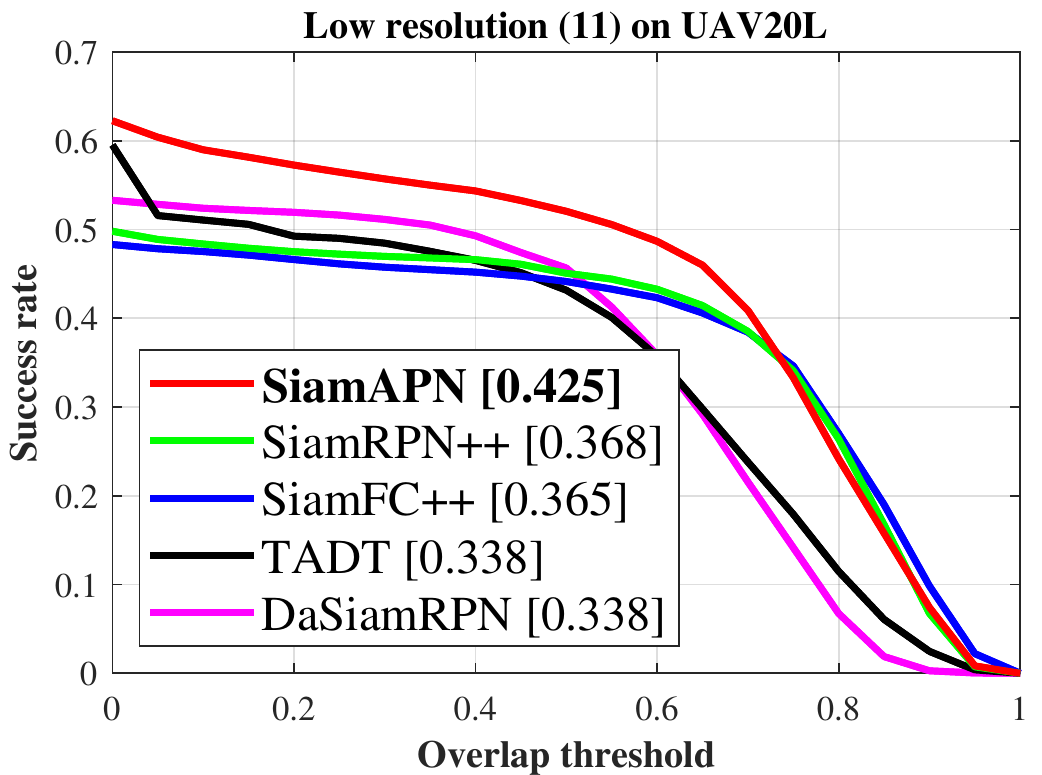}}
		
	\subfloat[Parital Occlusion on VisDrone2018-SOT-test (left) and UAV20L (right)]
	{\includegraphics[width=0.245\linewidth]{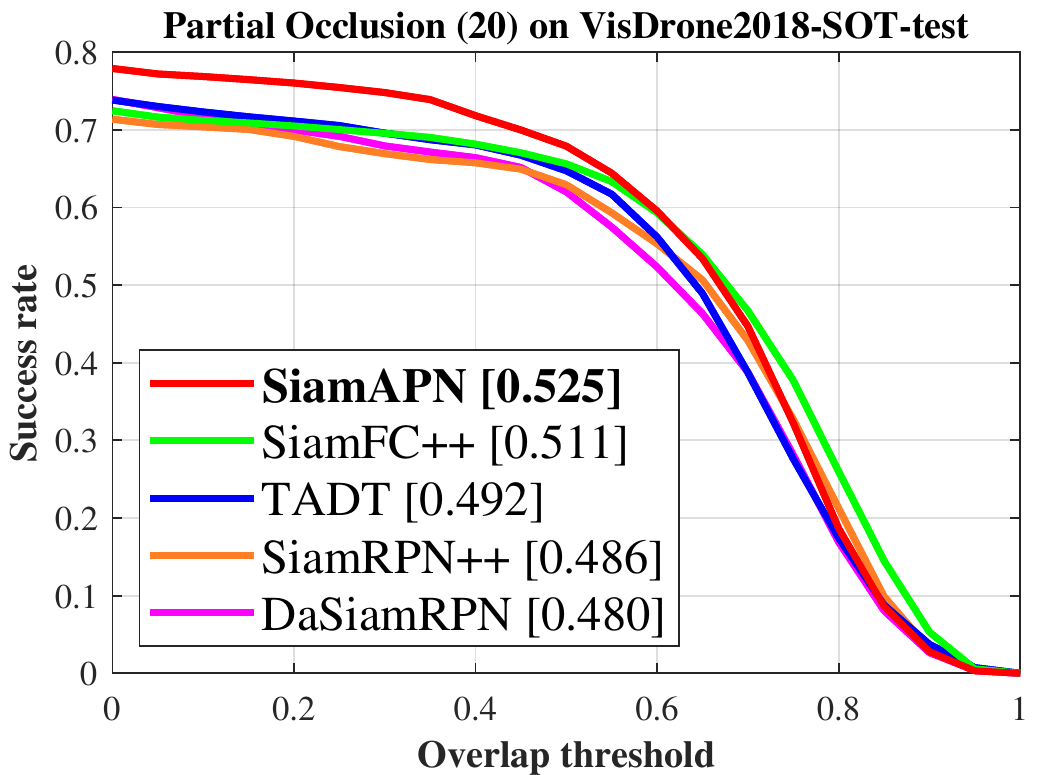}
	\includegraphics[width=0.245\linewidth]{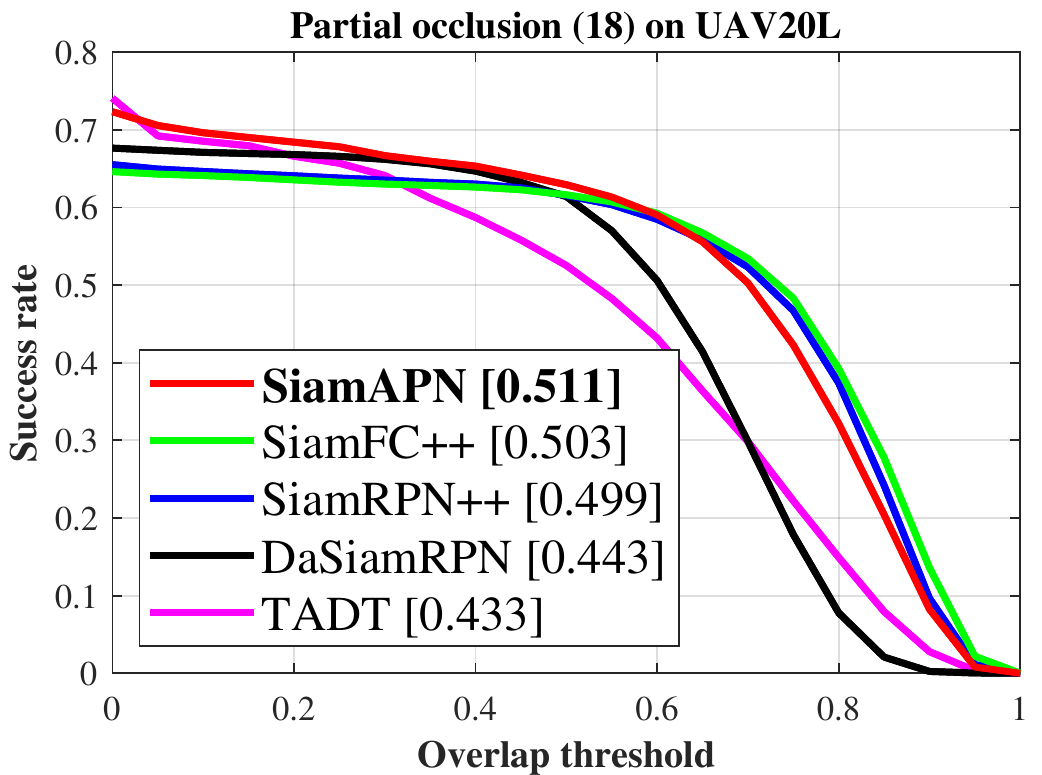}
	\label{fig:e}}
	\subfloat[Full Occlusion on VisDrone2018-SOT-test (left) and UAV20L (right)]
	{\includegraphics[width=0.245\linewidth]{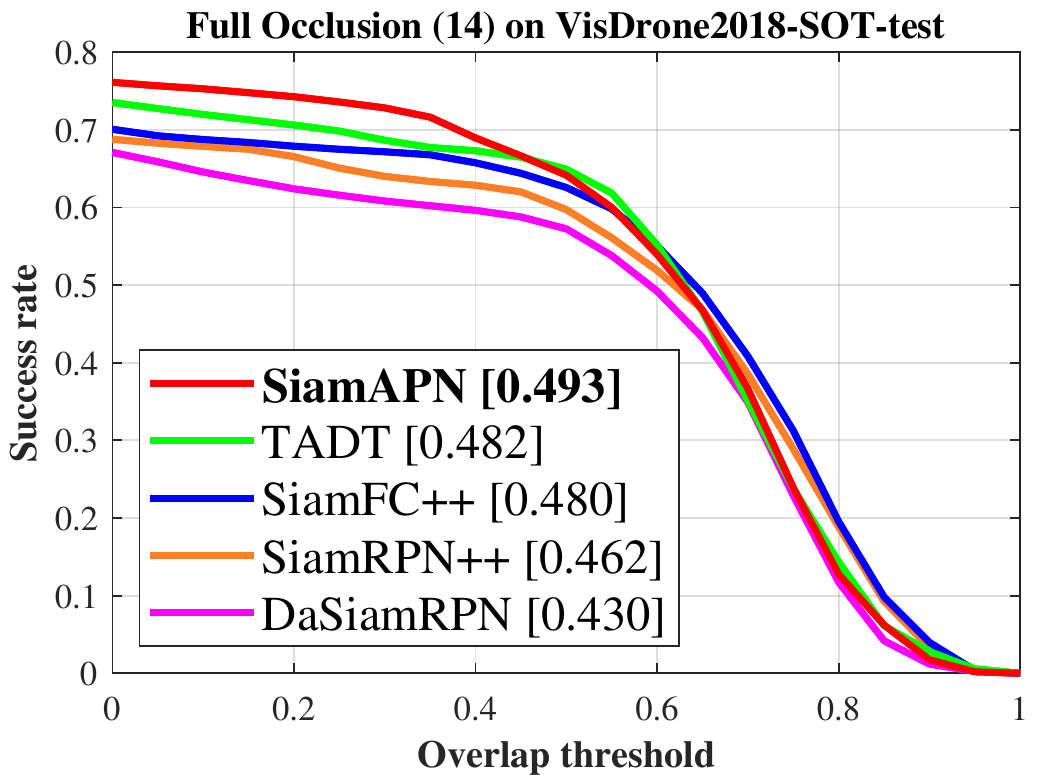}
	\includegraphics[width=0.245\linewidth]{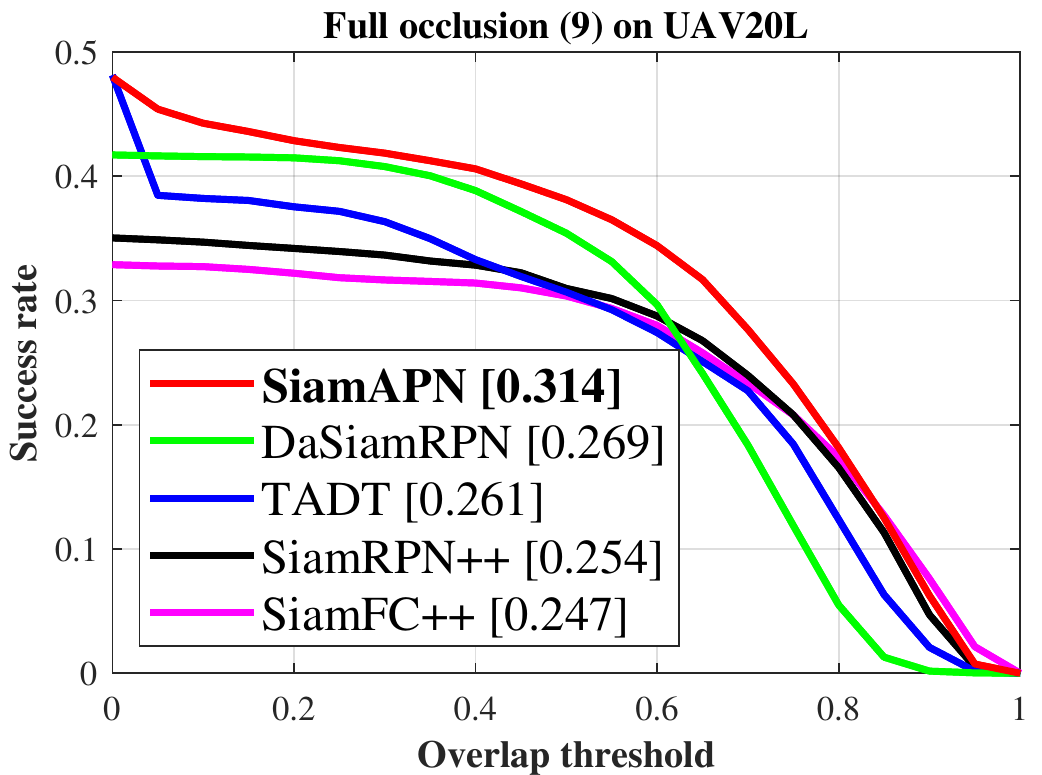}}

	\caption{Attribute-based comparison with top 5 trackers on four UAV-specific challenges, \textit{i.e.}, fast motion, low resolution, partial occlusion, and full occlusion. It shows that SiamAPN maintains robustness in different conditions.} 

	\label{fig:attribute}
\end{figure*}
%fast-motion, partical occlusion, full occlusion, and low resolution from UAV123@10fps~\cite{Mueller2016ECCV}, UAV20L~\cite{Mueller2016ECCV}, and VisDrone2018-SOT-test~\cite{wen2018visdrone}
\begin{table}[b]
	\footnotesize
	\setlength{\tabcolsep}{2.5mm}
	\centering

	\caption{Average precision as well as AUC score of top 5 trackers on three benchmarks. \textcolor[rgb]{ 1,  0,  0}{\textbf{Red}} and \textcolor[rgb]{ 0,  1,  0}{\textbf{green}} fonts indicate the first and second best results respectively.}
	\centering
	\setlength{\tabcolsep}{0.8mm}
	{
		\begin{tabular}{l c c c c c}
			\hline
			\hline
			{\textbf{Trackers}}&{\textbf{SiamAPN}}&{SiamFC++}&{SiamRPN++}&{DaSiamRPN}&{TADT}\\  
			\hline
			\textbf{Precision (\%)} & \textbf{\textcolor[rgb]{1  0  0}{76.2}} & {73.8} & \textbf{\textcolor[rgb]{0  1  0}{73.9}} & {69.2}&{68.5}
			\\ 
			\textbf{AUC (\%)} & \textbf{\textcolor[rgb]{1  0  0}{56.3}} & \textbf{\textcolor[rgb]{0  1  0}{56.2}} & {54.9} & {49.1}&{50.3}  \\
			\hline
			\hline
		\end{tabular}%
	}
	\label{tab:fps}%
\end{table}%
\subsection{Evaluation on UAV benchmarks}
\subsubsection{Overall performance}
The proposed method outperforms other state-of-the-art trackers on all three benchmarks.

\textbf{UAV20L:}
Consisting of 20 long-term sequences (2934 frames per sequence on average) with various challenging scenes including over 58k frames, UAV20L~\cite{Mueller2016ECCV} is designed especially for practical tracking. As shown in Fig.~\ref{fig:overperform}, SiamAPN performs favorably with an improvement of 2.5\% on precision and 0.6\% on AUC score against the second-best tracker.

\textbf{VisDrone2018-SOT-test:}
For evaluating the tracking algorithms in complex conditions, the VisDrone2018-SOT-test~\cite{wen2018visdrone} is created for UAV tracking with numerous visual challenges. As shown in Fig.~\ref{fig:overperform}, SiamAPN has outperformed all other trackers in terms of precision (0.814) and AUC (0.584), followed by UDT+ (0.786, 0.578).

\textbf{UAV123@10fps:}
Consisting of 123 sequences, UAV123@10fps~\cite{Mueller2016ECCV} is downsampled from a 30FPS version of sequences. Since the frame interval becomes larger, the challenges of many scenarios are increased, such as fast motion and low resolution. Consequently, to evaluate the ability of trackers in the face of fast motion comprehensively, UAV123@10fps~\cite{Mueller2016ECCV} is more appropriate than UAV123. The tracking performances of all trackers are presented in Fig.~\ref{fig:overperform}, SiamAPN ranks first in terms of precision (0.752), while maintaining the favorable performance on success rate (0.566).

The continuous competitive performance on the three benchmarks demonstrates the robustness of SiamAPN. To validate the top rank trackers in more detail, TABLE~\ref{tab:fps} presents the average precision and AUC of the top 5 trackers in three benchmarks, respectively. SiamAPN achieves the best performance in terms of precision and AUC score. Especially, SiamAPN has achieved an average improvement of 2.3\% in precision (0.762) compared to SiamRPN++ (0.739).
\begin{figure}[t]
	\centering
	\includegraphics[width=1\linewidth]{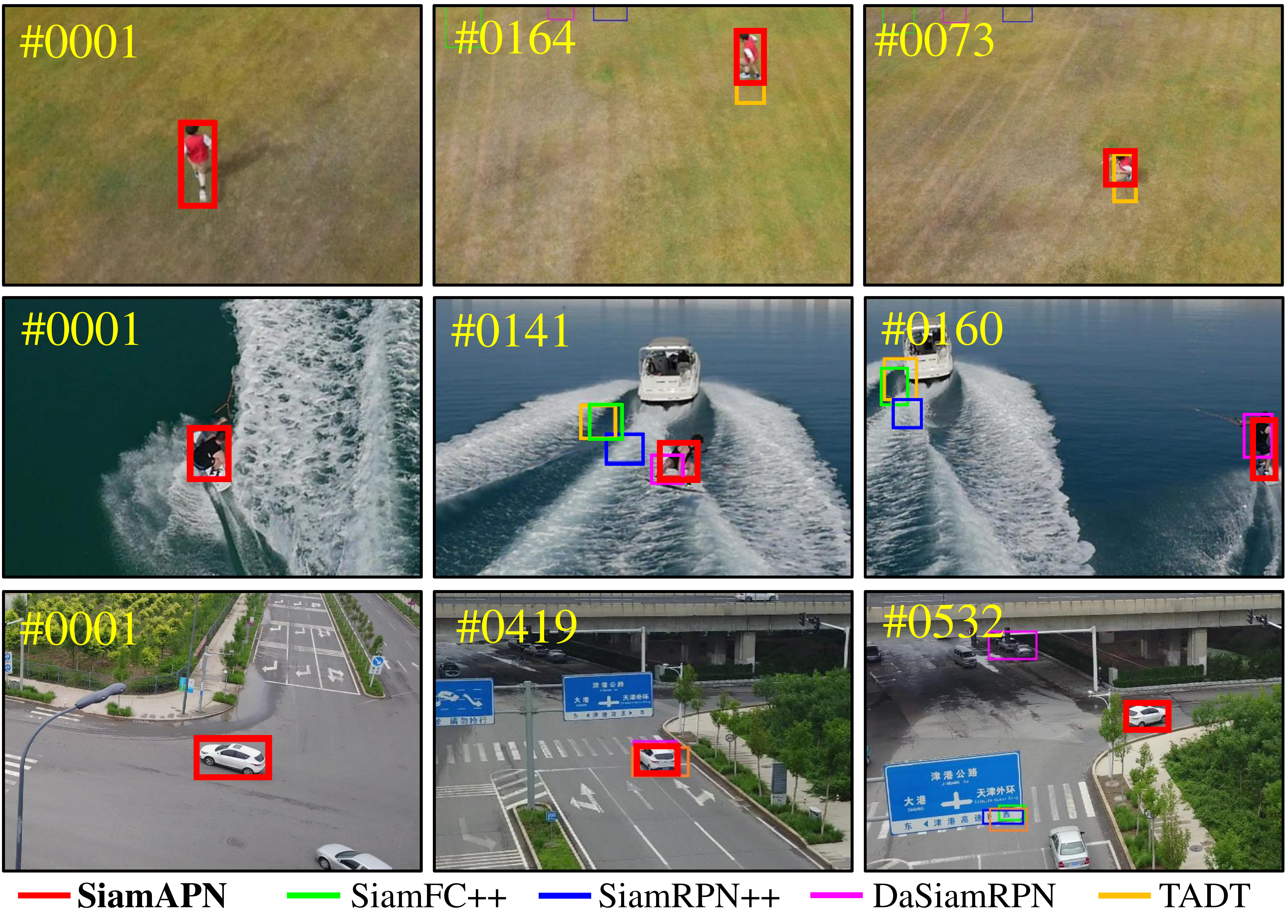}
	\vspace{-12pt}
	\caption{Screenshots of \textit{person7\_2}, \textit{wakeboard4} from UAV123@10fps, and \textit{uav0000180\_00050\_s} from VisDrone2018-SOT-test. More tracking videos can be found here: \url{https://youtu.be/tp3YIb2XHVk}.}
	\vspace{-20pt}
	\label{fig:5}
\end{figure}
\subsubsection{Attribute-based performance}
\label{subsec:attr}
Attribute-based evaluation results on UAV benchmarks are shown in Fig.~\ref{fig:attribute}. We compare the AUC score of the UAV-specific attributes of the top 5 trackers. Attributing to the flexible anchors generated by APN, SiamAPN can avoid neglecting objects in different scenes. 
In the fast motion scenarios, SiamAPN improves SiamFC++ by 3.9\% in the AUC score on average and improves SiamRPN++ by 5.1\% in the AUC score on average. In partial occlusion, full occlusion, and low resolution, SiamAPN has a superiority of 1.1\%, 4\%, and 3.9\% (AUC) on average respectively compared to SiamFC++ by utilizing flexible anchors. In addition, SiamAPN also outperforms significantly SiamRPN++ in terms of partial occlusion by 2.55\%, full occlusion by 4.55\%, and low resolution by 3.55\% on average. These cheerful results demonstrate the robustness of SiamAPN in various UAV tracking conditions. 

\subsubsection{Qualitative evaluations}
Some qualitative evaluations are released in Fig. \ref{fig:5}. It clearly illustrates that owing to the APN and feature fusion network, our SiamAPN tracker maintains superior accuracy and robustness under fast motion and severe occlusion scenes compared with other state-of-the-art trackers.
%\subsubsection{Key parameter analysis}
%To analyze the influence of the classification weight $\lambda_{cls1}$, $\lambda_{cls1}$ is set to different values for analyzing. Starting from 0.58, $\lambda_{cls1}$ increases in a small step of 0.1 until 1.78. When approaching the best point, the step size is further reduced to 0.02. As $\lambda_{cls1}$ increases, more attention is paid to the classification of adaptive anchors. Obviously, there is an optimum value for better classification. When $\lambda_{cls1}$ is set to 1.18, the AUC and precision reach the highest point shown in Fig.~\ref{fig:parameters}. However, as $\lambda_{cls1}$ continues to increase, too much attention is drawn to the APN classification branch, 
%\begin{figure}[t]
%	\centering
%	\includegraphics[width=1\linewidth]{para.pdf}
%	\vspace{-10pt}
%	\caption{Key parameter analysis of $\lambda_{cls1}$ on UAV123@10fps. When the $\lambda_{cls1}=1.18$, SiamAPN achieves the best overall performance.}
%	\vspace{-10pt}
%	\label{fig:parameters}
%\end{figure}
%the information in the other two branches is not assigned enough weight to the performance. 
%Therefore, when $\lambda_{cls1}$ is set to an appropriate value, APN can indeed favorably boost the tracking performance. Consequently, $\lambda_{cls1}=1.18$ is chosen for the best performance.

%%%%%%%%%%%%%%%%%%%%%%%%%%%%%%%%%%%%%%%%%%%%%%%%%%%%%%%%%%%%%%%%
%%%%%%%%%%%%%%%%%%%%% Section 5: CONCLUSIONS %%%%%%%%%%%%%%%%%%%
%%%%%%%%%%%%%%%%%%%%%%%%%%%%%%%%%%%%%%%%%%%%%%%%%%%%%%%%%%%%%%%%

\section{Conclusion}\label{sec:CONCLUSIONS}
In this work, for overcoming the special challenges in aerial tracking while maintaining promising efficiency, a novel two-stage tracking approach is proposed, which combines the two mainstream Siamese-based methods. The adaptive anchors introduced by APN eliminate many hyper-parameters and alleviate the problem of imbalanced samples. Merely adopting the APN can decrease the difficulty of classification while increasing those in regression. Therefore, to avoid the influence on regression, the feature fusion network is conducted. By virtue of the feature fusion, the location information of anchors is incorporated, enhancing the semantic knowledge of feature maps, which is essential for robust aerial tracking. By exploiting the multi-classification structure, the classification accuracy is also significantly boosted, improving the discriminability of SiamAPN under complex conditions. Comprehensive experiments on three authoritative benchmarks have strongly demonstrated the competitive performance of SiamAPN. Consequently, we are convinced that our work can promote the development of aerial tracking-related applications.
%%%%%%%%%%%%%%%%%%%%%%%%%%%%%%%%%%%%%%%%%%%%%%%%%%%%%%%%%%%%%%%%%%%%%%%%%%%%%%%%

%%%%%%%%%%%%%%%%%%%%%%%%%%%%%%%%%%%%%%%%%%%%%%%%%%%%%%%%%%%%%%%%%%%%%%%%%%%%%%%%
\vspace{-5pt}
\section*{Acknowledgment}
This work is supported by the National Natural Science Foundation of China (No. 61806148) and the Natural Science Foundation of Shanghai (No. 20ZR1460100).

%%%%%%%%%%%%%%%%%%%%%%%%%%%%%%%%%%%%%%%%%%%%%%%%%%%%%%%%%%%%%%%%%%%%%%%%%%%%%%%%

\bibliographystyle{IEEEtran}  %这是你要使用的格式,比如要投IEEE,就写IEEEtran
\bibliography{IEEEabrv,ref}%这个是加载你的bib,你可以理解从文献数据库中加载要引用的文献

\end{document}